\documentclass[review]{elsarticle}

\usepackage{lineno,hyperref}
\modulolinenumbers[5]

\usepackage{csquotes} 
\usepackage{amssymb,amsmath} 
\usepackage{subfigure} 
\usepackage{float} 
\usepackage{booktabs} 
\usepackage{multirow} 
\usepackage[table,xcdraw]{xcolor} 
\usepackage{multicol}
\usepackage{graphicx}
\usepackage{dsfont}
\usepackage{tipa} 

\usepackage{caption}

\usepackage{scalerel,stackengine}
\stackMath
\newcommand\reallywidehat[1]{%
\savestack{\tmpbox}{\stretchto{%
  \scaleto{%
    \scalerel*[\widthof{\ensuremath{#1}}]{\kern-.6pt\bigwedge\kern-.6pt}%
    {\rule[-\textheight/2]{1ex}{\textheight}}
  }{\textheight}%
}{0.5ex}}%
\stackon[1pt]{#1}{\tmpbox}%
}

\DeclareMathOperator*{\argmin}{arg\,min}

\usepackage{hyperref}
\usepackage{breqn}

\usepackage{flexisym} 

\usepackage{mathtools} 
\usepackage{color} 

\journal{Pattern Recognition - Journal - Elsevier}

\begin{document}

\begin{frontmatter}

\title{\textbf{ADS-ME}: Anomaly Detection System for Micro-expression Spotting}

\author[mymainaddress]{Dawood Al Chanti\corref{mycorrespondingauthor}}
\cortext[mycorrespondingauthor]{Corresponding author}
\ead{dawood.alchanti@gmail.com}

\author[mymainaddress]{Alice Caplier}
\ead{alice.caplier@gipsa-lab.grenoble-inp.fr}

\address[mymainaddress]{Univ. Grenoble Alpes, CNRS, Grenoble INP\fnref{footnote1}, GIPSA-lab, Image and Signal Processing Department, 38000 Grenoble, France.}
\fntext[footnote1]{Institut polytechnique de Grenoble.}

\begin{abstract}

Micro-expressions~(MEs) are infrequent and uncontrollable facial events that can highlight emotional deception and appear in a high-stakes environment. This paper propose an algorithm for spatiotemporal MEs spotting. Since MEs are unusual events, we treat them as abnormal patterns that diverge from expected Normal Facial Behaviour~(NFBs)~patterns. NFBs correspond to facial muscle activations, eye blink/gaze events and mouth opening/closing movements that are all facial deformation but not MEs. We propose a probabilistic model to estimate the probability density function that models the spatiotemporal distributions of NFBs patterns. To rank the outputs, we compute the negative log-likelihood and we developed an adaptive thresholding technique to identify MEs from NFBs. While working only with NFBs data, the main challenge is to capture intrinsic spatiotemoral features, hence we design a recurrent convolutional autoencoder for feature representation. Finally, we show that our system is superior to previous works for MEs spotting.

\end{abstract}

\begin{keyword}
Micro expressions, Anomaly detection, Recurrent Convolutional Autoencoder, Density estimation.
\end{keyword}

\end{frontmatter}


\section{Introduction}
\noindent A face is a natural nonverbal channel that conveys a number of essential social signals. It mediates facial communicative cues, that is, facial expressions, to communicate emotions even before people verbalize their feelings. Most of the existing automated systems for facial expression analysis attempt to recognize a prototypic emotional macro-expression such as angry, disgust, fear, happiness, sadness, and surprise. These macro-expressions sometimes pretend the genuine emotions. On the contrary, facial Micro-Expressions (MEs) are repressed, they are involuntary expressions that appear when people tries to mask their true emotion. They tend to be more probable in a high-stakes situation as showing emotions is risky.

Though being very subtle, MEs contain information about the true emotion \cite{ekman2009lie}, therefore automating the process of spotting and classifying them is desirable. MEs are a promising cue as it has a wide range of applications such as affect monitoring \cite{porter2008reading}, lie detection \cite{bernstein2009tell}, and clinical diagnosis \cite{russell2006pilot}.

Micro-expressions are characterized by a sparse activation of subtle facial movements, a brief duration (1/25 to 1/3 second) and a fast motion \cite{ekman2009lie},\cite{warren2009detecting}. Visual reading of MEs by experts is only around 45\% \cite{endres2009micro},\cite{frank2009see}. Obviously, spotting MEs is a challenging problem, as there is a need for more descriptive facial feature displacements and motion information. Accordingly, high-speed camera is a must to capture the speed and subtlety of MEs. But the usage of such camera tends to produce noisy data wherein eye-related events (blinking of the eyes, changes in gaze direction) and facial muscle activations are reinforced and then can be confused with MEs.

\begin{figure}[!t]
\centerline{\includegraphics[width= 5 in]{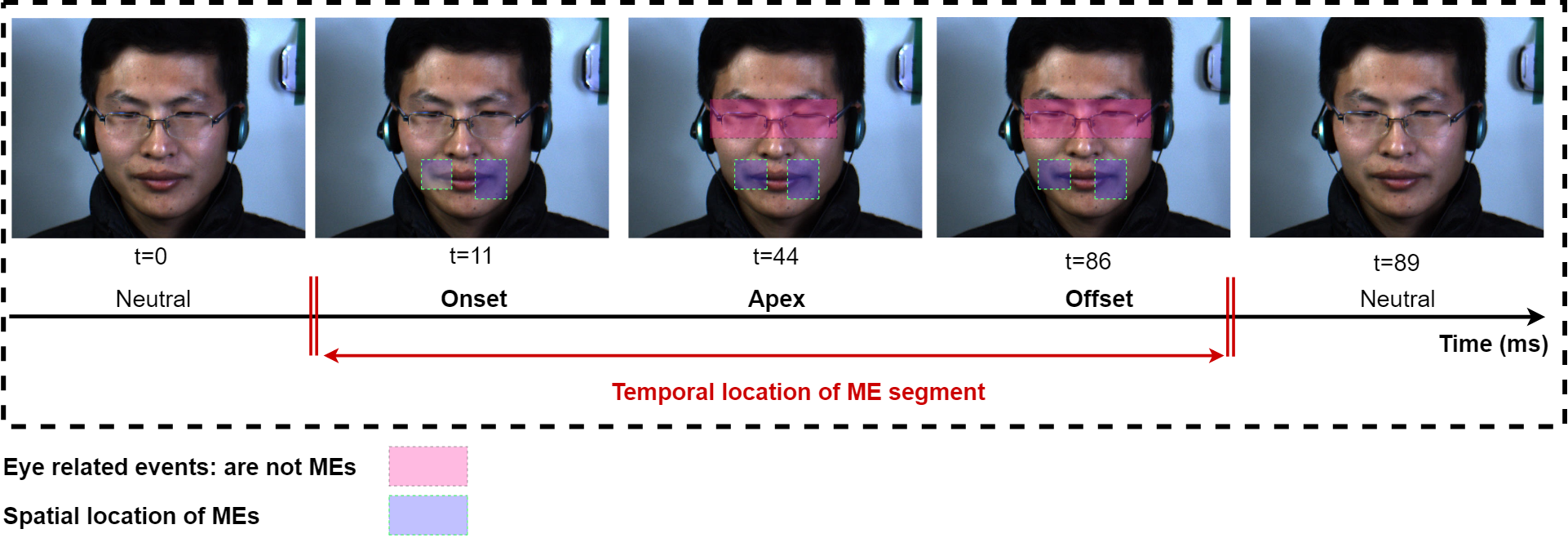}}
\caption{Spatio-temporal MEs detection. A face with a micro-expressions that appear around the lip corners (spatial location) associated with a fast eye blinking at the same time.}
\label{MEDetection}
\end{figure}

Researches on MEs analysis proceed mainly along two dimensions: (1) \emph{MEs Spotting} for localizing the temporal occurrence of MEs and (2) \emph{MEs Recognition} for determining the category of the emotional state. MEs Recognition study has received more attention while assuming that MEs segment have already been localized. Conversely, few studies have been reported regarding the problem of MEs spotting, though being the primary step for MEs Recognition. In this paper, the goal is put on building a model that:

\begin{enumerate}
    \item Accurately detects the temporal location (onset-offset frames) of MEs.
    
    \item Determine the spatial location (pinpointing the facial region/regions) of the subtle facial deformations involved in MEs.
    
    \item Effectively deals with parasitic movements (e.g. eye-related events) as shown in figure~\ref{MEDetection}. 

\end{enumerate}


To achieve our goal, we reformulate the problem of MEs spotting into a problem of \emph{Anomaly Detection} and we propose an Anomaly Detection System for Micro-expression Spotting, referred to \textbf{ADS-ME}. In such manner, anomaly refers to the unusual pattern occurring irregularly or being different from other usual normal patterns. Herein, we assigned MEs to anomaly class as it represent abnormal facial behaviour and for the normal class we consider everyday life events (macro-expression, speech, shouting,...)~as Natural Facial Behaviours~(NFBs). NFBs events consist of fast blinking, eye-gaze changes, facial action unit activations, global head movements and mouth movements (opening/closing). The anomaly class (MEs) is often absent during training, poorly sampled or not well defined while the normal class (NFBs) is well characterized and have a lot of normal samples in the training data. In this study, anomalous behaviors are appointed to MEs due to two main reasons:

\begin{enumerate}
    \item ME events occur infrequently in comparison to NFBs events, and they are not well represented within the image sequences available for modeling.
    \item ME events exhibit significantly distinctive spatio-temporal information with respect to NFBs events.
\end{enumerate}


Due to the lack of data of the anomaly class, designing a statistical approach for modeling the normal class distribution and rejecting samples not following this distribution is not straightforward. Therefore, we propose a decoupled process. First, to accurately chart the intrinsic spatiotemporal links of the normal class, a Recurrent Convolutional AutoEncoder (RCAE) is build and learned to entangle the different explanatory factors of spatiotemporal variations in the normal class. Then, the density parameters of the normal class are estimated in the new subspace (latent features from RCAE) using a Gaussian Mixture Model (GMM). Finally, the weighted log-likelihood is computed for ranking the output at each time instance. Typically, a low probability score is expected for the anomaly class as it does not belong to the modeled normal behaviour distribution. Identifying MEs from NFBs based on the obtained likelihood requires thresholding, and to do so we propose an adaptive thresholding technique.


Since subtle expressions occur in highly localized regions of all the face and across time, we consider a spatiotemporal anomalous sample to be a region, wherein the data values within the spatiotemporal region of MEs are different from the ones in its neighborhoods. To provide spatial localization, an image is partitioned into equal regions while preserving their temporal links and a probability distribution that represents each region is estimated. Determining the spatial location has the advantage of masking unrelated facial movements for further analysis and to enhance MEs classification. Partitioning the image into blocks has the advantage of sampling a large number of individual blocks which provide a greater statistical power while estimating the probability density function (PDF). Moreover processing small spatial regions is computationally less expensive.

Our approach has the benefit of requiring only training events from the normal class and not requiring any data from anomalous events. But due to the fact that MEs databases are acquired using high-speed cameras, this tends to produce noisy data, such as eye-related events or head motion for instance. Such normal class events are poorly represented in the training data in the same manner MEs are poorly represented. This might lead to confusion of such events with those occurring with an ME. However, since our interest is to look after modelling the normal class distribution, establishing a good amount of data that represent poorly sampled normal class event is needed. To overcome this challenge, we propose a temporal sampling in which each block sequence is divided into multiple temporal segments at varying temporal resolution thus defining \textbf{instances}. A collection of instances is represented as a \textbf{bag}. The whole bags that correspond to NFBs are used to train RCAE. Then, a probability density function for each bag is separately estimated. At inference time, the full video clips including normal and abnormal patterns are presented. Then for each block at a time instance, the weighted log-likelihood is computed and followed by adaptive thresholding for MEs spotting.

In summary, this paper makes the following contributions:
\begin{itemize}
    \item A Spatio-temporal algorithm for spotting ME segments including the onset and offset frames while spatially localizing in each frame the regions involved in the ME process.
    \item A deep learning approach to capture facial spatial features and motion changes of NFBs.
    \item The formulation of MEs detection as an anomaly detection process and a statistical model for estimating the PDF of normal facial behaviours while associating a discriminating score to spot MEs.
    \item  A grid pattern based method alongside a temporal multiscaling to sub-sample local changes at varying temporal resolutions for empowering the PDF estimation.
    \item An adaptive thresholding technique for identifying MEs from NFBs.
\end{itemize}

The structure of this paper is as follows: Section \ref{RW} discusses the related work, Section \ref{algo} explains the detection algorithm in detail, Section \ref{exp} presents the used MEs databases and the obtained performances and Section \ref{conc} concludes the paper.

\section{Related Work}
\label{RW}





\noindent \textbf{Micro-expressions.} Recent studies towards enhancing MEs detection have been reported. Pfister et al.~\cite{pfister2011recognising} developed an innovative framework in 2011, where a Temporal Interpolation Model (TIM) alongside multiple kernel learning (MKL) to recognize short expressions is developed. The authors showed that TIM is beneficial while using standard camera of 25 frames per second (fps) so that it can help matching the detection accuracy as that of 100 fps. To address large variations in the spatial appearances of MEs, the face geometry is cropped and normalized according to the eye positions from a Haar eye detector and the feature points from an Active Shape Model (ASM). The 68 ASM feature points are transformed using Local Weighted Mean transformation to a model face followed by a spatiotemporal feature extraction using a hand-crafted descriptor, mainly the LBP-TOP. MKL and random forest are utilized to classify MEs from non-MEs. 

In 2014, a number of algorithms start to appear. A weighted feature extraction scheme has been proposed by Liong et al.~\cite{liong2014subtle} to capture subtle micro-expressions movements based on Optical Strain. It is defined as the relative amount of deformation of an object. Its ability to capture muscular movements on faces within a time interval makes it suitable for MEs research. Contrary to the Optical Flow which is highly sensitive to any changes in brightness. The motion information is derived from optical strain magnitudes and used as a weighting function for the LBP-TOP feature extractor. The last step directly uses the motion information in order to avoid the loss of essential information from its original image intensity values. Then, an SVM is used for classification.

A training free based method for automatically spotting rapid facial movements is proposed by Pietikainen et al.~\cite{moilanen2014spotting}. The method relies on analyzing differences in appearance-based features of sequential frames. It aims at finding the temporal locations and to provide the spatial information about the facial movements. It is mainly composed of five steps:~(1) tracking stable facial points followed by image alignment;~(2) dividing frames into blocks;~(3) extracting local features using LBP descriptor;~(4) calculating the \textchi$^{2}$ distance within a defined time interval for each block of sequential frames;~(5) handling the difference matrix by:~(i) obtaining difference values for each frame by averaging the highest block difference values,~(ii) contrasting relevant peaks by subtracting the average of the surrounding frames' difference values from each peak, and~(iii) using thresholding and peak detection to spot rapid facial movements in the video. The authors showed through their experimental analysis that the proposed method is sensitive to detect other facial events such as eye blinks, global head movements or brightness variation that are not produced by MEs.

To analyze MEs, Yan et al.~\cite{yan2014quantifying} assess the spatiotemporal representation. The authors defined a Constraint Local Model algorithm to detect faces and track feature points. Based on these points, the ROIs on the face are drawn. Then, the LBP descriptor is used for feature extraction from the defined ROIs and mainly for texture description. Finally, the rate of texture change is obtained by computing the difference between the first frame and the other frames.

The first algorithm that has spotted the onset and the offset frames of MEs was proposed by Patel et al.~\cite{patel2015spatiotemporal} in 2015. The authors compute the optical flow vector around facial landmarks and integrate them in local spatiotemporal regions. A heuristics to filter non-micro expressions is introduced to find the appropriate onset and offset times. Finally, false detections as head movements, eye blinks and eye gaze changes are reduced by thresholding.

Xia et al.~\cite{xia2016spontaneous} in 2016 highlighted the main problems of detecting micro-expressions such as subtle head movements and unconstrained lighting conditions. To face these challenges, a random walk model is introduced to calculate the probability of individual frames being MEs. Then an Adaboost model is utilized to estimate the initial probability for each frame and the correlation between frames is considered into the random walk model. The ASM and Procrustes analysis are used to describe the geometric shape of a human face. Finally, the geometric deformation is modeled and used for Ada-boost training.

The detection precision in the latest works since 2017-up-to-date increased. Borza et al.~\cite{borza2017high} proposed a framework which can detect the frames in which MEs occur as well as determine the type of the emerged expression. Their method uses motion descriptors based on absolute image differences. Ada-boost is used to differentiate MEs from non-MEs. The facial ROI is restricted to ten facial regions in which the 68 facial landmarks reside. Li et al. \cite{li2017towards} proposed to spot MEs using feature difference contrast and peak detection. Their method starts by dividing the facial region into equal size blocks and then track each block along the sequence. Spatiotemporal feature extraction using the LBP-TOP descriptor over each block is utilized and then followed by feature difference analysis and thresholding. Borza et al.~\cite{borza2017real} capture the movements in facial regions based on an absolute difference technique with random forest as classifer. Duque et al. \cite{duque2018micro} spotted MEs in a video by analyzing the phase variations between frames obtained from a Riesz Pyramid. This method is capable at differentiating MEs from eye movements. Lastly, Li et al.~\cite{8373893} proposed to improve the detection accuracy by recognizing local and temporal patterns of facial movements. The method consists of three parts, a pre-processing step to detect facial landmarks and extract the ROIs, then the extraction of local temporal patterns from a projection in PCA space and eventually detecting of MEs using an SVM classification.

Our approach shares some similarities with previous works regarding the way to pre-process the facial image. The face is firstly detected, cropped and followed by block division for partitioning the image space. Our proposal does not require any face alignment since our algorithm does not depend on feature difference analysis, and therefore no further tracking algorithm is needed. Regarding spatiotemporal feature extraction, among many local spatiotemporal descriptors, the literature has focused on using LBP-TOP and 3D Histogram of Oriented Gradient~(3D-HOG). Here, we decide to take advantage of deep learning based methods to provide a spatiotemporal representation that is robust against background changes, illumination and various environmental changes. Finally, the literature utilizes the extracted features to capture the dissimilarities between MEs and non-MEs frames based on either a classification method or a training free based method. Here, we propose a probabilistic framework based on GMM and adaptive thresholding to identify MEs from non-MEs.




\noindent \textbf{Anomaly detection.} This part introduces preliminaries and reviews the state of the art in video anomaly detection based on recent surveys \cite{chandola2009anomaly}, \cite{sodemann2012review} and \cite{kiran2018overview}. Anomaly detection is the process of identifying abnormal patterns that correspond to changes in appearance and motion. Typically abnormal events occur rarely and are hardly to annotate or not sufficiently represented. In such a case, class imbalance problems can occur because of the divergence between normal and abnormal sample ratios. Consequently, when only normal behavior samples are easily accessible, it is possible to utilize a semi-supervised method, which only uses normal data to build the model. Existing approaches for semi-supervised methods in the literature can be roughly placed into two categories:~(i)~\emph{Lossless compression/Reconstruction based methods}: the main principle of these methods comes from information theory perspectives \cite{6310873} in measuring the information quantity, and detects anomalies according to compression result instead of statistics. These methods assume that anomalies cannot be effectively reconstructed from low-dimensional projections and therefore anomalies will result from higher reconstruction error. However, due to the complex structures of normal class (images and videos) and in some domains due to the fine-granularity of spatial and motion information, obtaining an accurate normal class data without containing few abnormal data is not an easy task. The lurk of abnormal data into normal class data might generate indistinguishable reconstruction error and thus limit the performance of such methods. Recent anomaly detection models have utilized deep learning methods using AutoEncoder topology \cite{marchi2015novel}, \cite{yang2015unsupervised},\cite{marchi2015non} and the reconstruction error is used as an activation signal to detect anomalies. The performances of these models are promising but also report significant false positive rates; (ii)~\emph{Statistical models}: these models rely on data being generated from a particular distribution. Statistical anomaly detection techniques assume that normal data instances occur in the high probability regions of a stochastic model, while anomalies occur in the low probability regions of the stochastic model. These methods tend to fit a model to the given normal behavior data and then apply a statistical inference test to determine if an unseen instance belongs to this model or not. Instances that have a low probability of being generated from the learned model are declared as anomalies. Gaussian Mixture Models for anomaly detection are frequently used \cite{basharat2008learning}, \cite{laxhammar2009anomaly}, \cite{nikisins2018effectiveness} and \cite{lim2019heartrate} and show good performances. Such techniques assume that the data is generated from a Gaussian distribution. A threshold is applied to the anomaly scores to determine the anomalies. The main limitation of these methods lies down in its difficulty to directly address the curse of dimensionality problem due to multi- or high-dimensional data \cite{chandola2009anomaly}. 

To this end, we propose a recurrent convolutional autoencoder (RCAE) that is capable at preserving essential spatiotemporal information while generating a low-dimensional representation. Then, a statistical model is designed, by feeding the extracted spatiotemporal features into a Gaussian Mixture Model.

\section{Micro Expressions Detection Algorithm}
\label{algo}

\begin{figure}[!t]
\centerline{\includegraphics[width= 4.5 in]{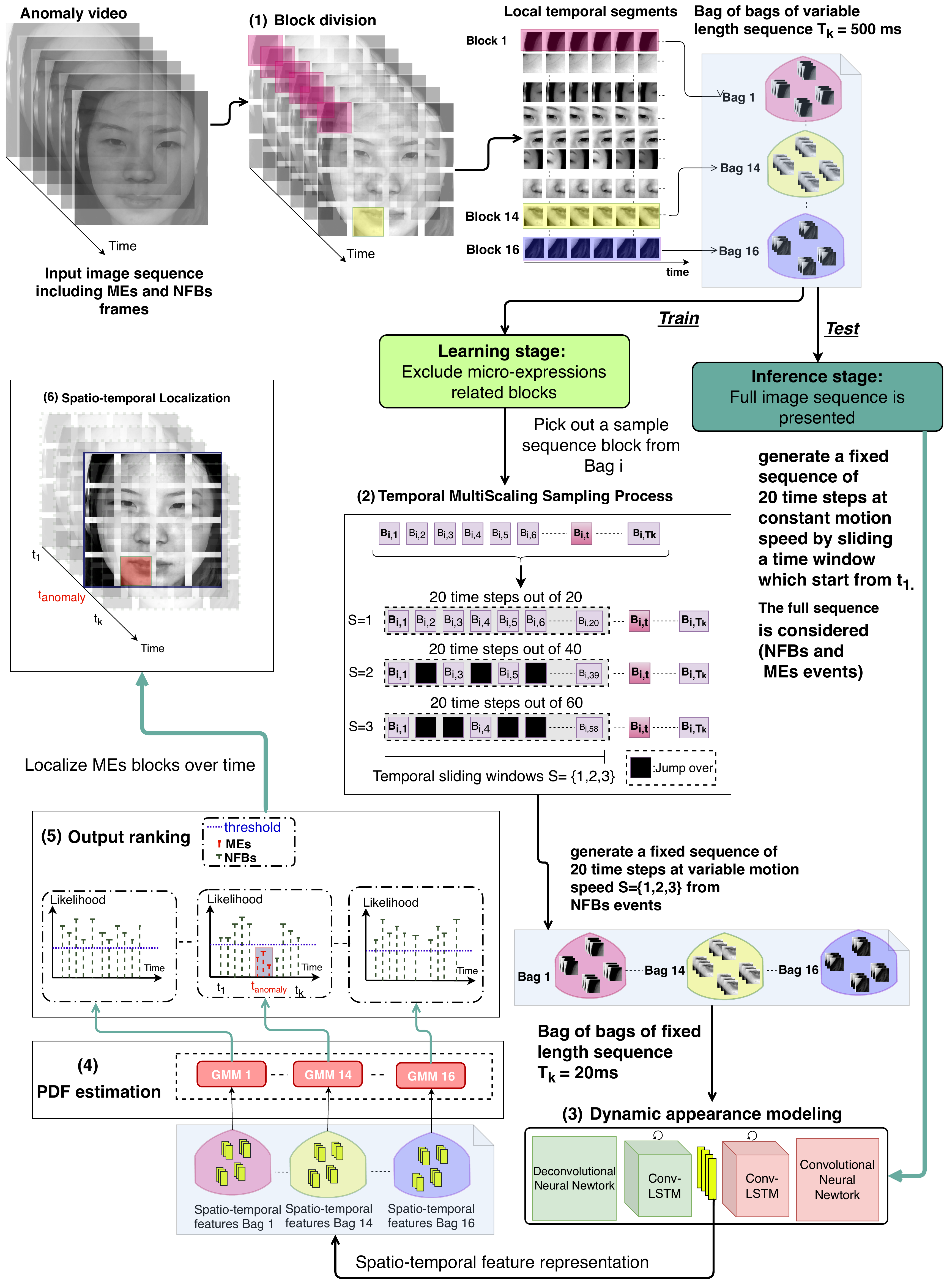}}
\caption{General block diagram of ADS-ME framework. Best seen in color.}
\label{generalblockdiagram}
\end{figure}

\noindent Figure~\ref{generalblockdiagram} shows a block diagram which explain how ADS-ME work for spotting and localizing MEs. Let us consider an anomaly video composed of $k$ frames that include NFBs and MEs; After pre-processing each frame, block division (step 1) is applied; The temporal link of each block is kept and a bags of block sequences is established with length $T_{k}$; During the \emph{Learning Stage}, blocks related to MEs sequence are suspended and only NFBs block sequences are considered. Within this stage, Temporal multiscaling (step 2) is applied and its output is what we refer to as \textbf{instances}: a sequence of block of fixed length $T_{20}$ with various motion speed; A collection of instances are used to train a RCAE for dynamic appearance modeling (step 3); The obtained spatiotemporal vectors of each bag are used to train a separate GMM model (step 4) that establish the regional distribution of each bag; During the \emph{Inference Stage}, the full video clip is considered and no temporal multiscaling is taken into account; However, a sliding window that covers 20 time step is used to generate a fixed length sequence in which its spatiotemporal features are encoded and fed to the estimated PDF to rank the output (step 5); An adaptive threshold is computed based on the input video; Afterwards deviated patterns (MEs) are spotted in time and space by applying the computed threshold over the weighted log-likelihood scores (step 6).

\subsection{The Learning Stage}
\label{learningstage}

\noindent 
A two processing steps are carried out, one to partition the face based on spatial grid and the other to sub-sample at various temporal speeds local NFBs samples via temporal multiscaling. Then, a two learning steps are executed separately: feature learning and PDF estimation. The aim is to model the NFBs distribution for each bag and to maximize its likelihood.

\subsubsection{Grid Pattern Based Method for Block Division}
\noindent The intuition behind grid pattern based methods is to divide every frame into blocks. The spatial connection between blocks is not considered but the temporal connection is kept. Then, for each block separately, a model is built to look for abnormal patterns that deviate from normal ones. By evaluating each block alone, the localization is possible. In this work, first the facial region is extracted using the Viola and Jones algorithm, then the face is re-sized to have a spatial resolution of $360 \times 360$ pixels. Afterwards, each frame is divided into $4\times4$ blocks. Hence, 16 local regions are obtained where individual block have a spatial resolution of $90 \times 90$ pixels. Blocks are separately tracked through time to keep the temporal information and considered as an observed data. By that, a large number of local spatial deformations are reachable. 

\subsubsection{Temporal Multiscaling for Sub-sampling}

\noindent Analysing subtle changes in facial behavior is a challenging problem because of its fine-granularity. MEs in high spatiotemporal resolution video may simultaneously co-occur with some NFBs events mainly eye-related events. Consequently, such NFBs events that co-occur with MEs are also infrequent, even though they are not MEs events. Presenting insufficient normal data samples might lead to normal data outliers. Obtaining sufficient information to distinguish NFBs from MEs is critical. Transforming outliers to inliers requires generating a good amount of training sequences. Therefore, we propose a temporal multiscaling method that generates different dynamic motion information at various speeds along the temporal axis. It works by skipping some blocks frames at different time scales. It is targeted to generate a new fixed length sequence from a variable length sequence at various temporal scales. Obtaining fine motion information for normal data class helps to empower the feature extraction process while learning RCAE. It yields to a distinctive spatiotemporal feature representation, wherein fine space-motion that represents NFBs events can be distinguished from those related to MEs events.

Our proposal for Temporal Multiscaling Sampling (\emph{TMS}) works by sliding a Temporal Window \emph{TW} that covers 20 time steps out of a variable length video $T_{k}$ as shown in figure~\ref{generalblockdiagram}. The main characteristic of a TW is that it can jump between frames according to a parameter $S=\{1,2,3\}$. With $S=\{1\}$, consecutive block frames are considered that are every 20 milliseconds (ms). With $S=\{2\}$, 20 ms out of 40 ms are covered with a jump of one between blocks. And with $S=\{3\}$, 20 ms out of 60 ms are covered with a jump of two between blocks. TW does not need to start from $t_{1}$. It starts randomly at any time instance such that it satisfies the condition of covering 20 ms (being the minimal duration of a ME). The optimal length of a TW is a hyper-parameter that is tuned via experimental setup. The output sequence after the TMS process has a shape of $20\times90\times90\times1$, 20 being the number of images in a sequence, 90 being the height and width of each block and 1 refers to the gray channel. The minimal number of data samples being generated after TMS process is $3\times16\times \text{n}$, where 3 represents the number of sliding windows performed over each sequence, 16 represents the number of blocks generated from each static image and $n$ is the original number of training sequences. To this end, another advantage of the TMS process is its ability to generate a huge number of samples with various deformations within image space and along the temporal axis which empowers the learning process of spatiotemporal feature extraction and statistical modeling.

\subsubsection{Spatiotemporal Feature Learning}
\begin{figure*}[!t]
\centerline{\includegraphics[width= 5 in]{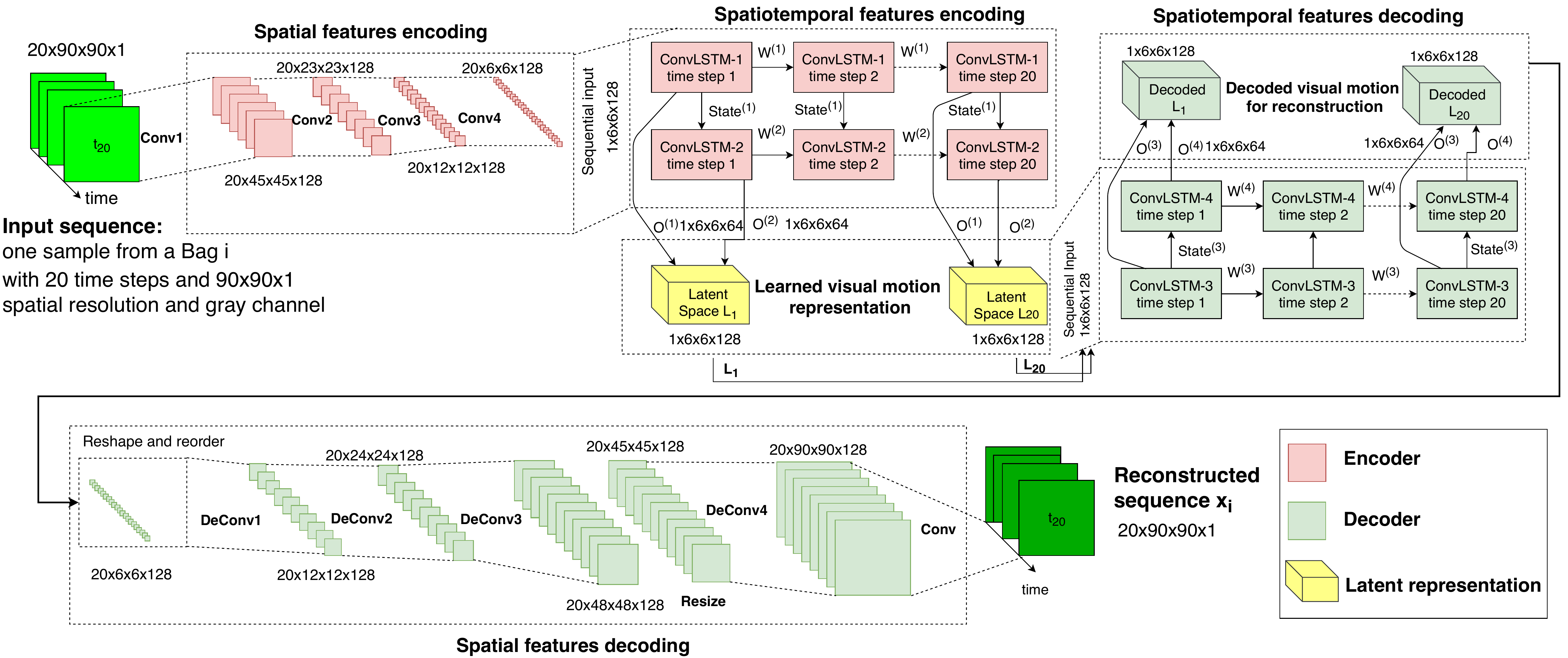}}
\caption{Visual motion modeling for normal facial behavior analysis using RCAE. Both the \textbf{Encoder} and the \textbf{Decoder} are made up of multilayered CNNs and ConvLSTMs.}
\label{RCAEblock}
\end{figure*}

\noindent NFBs and MEs are characterized by some distinctive spatial structure and temporal links between prominent facial regions over time. The temporal evolutions and spatial displacements allow to analyze the current situation relatively to the past, which is critical for describing the entire content of the input sequence. In this study, we decide to leverage a learning method to learn visual motion features related to NFBs over short-term and long-term temporal horizons without any annotation constraints. For this task, a Recurrent Convolutional Autoencoder (RCAE) is developed as shown in figure~\ref{RCAEblock}, wherein a convolutional network followed by a multilayer Convolutional Long Short-Term Memory (ConvLSTM) cell \cite{xingjian2015convolutional} is designed. It aims firstly at encoding the input sequence into a fixed length effective representation using a non-linear transformation. Secondly, it aims at extracting the spatial deformations and motion information. And more importantly, at memorizing the past states of spatial block motions by continuous updates of the cell states of the encoder. Afterwards, a decoder that has approximately the mirror architecture than the encoder is designed to map back the extracted spatiotemporal information into its original input space. The mean squared error is used as objective function. 


Given a set of training samples $\textbf{X}^{\textbf{train}}$ with NFBs blocks only, the main goal is to learn a feature representation (referred as \emph{latent representation} \textbf{L}) that captures normal behavior spatiotemporal patterns. Let $x_{i} \in \mathbb{R}^{20 \times 90 \times 90 \times 1}$ be a sample coming from a facial sub-region block \textit{i} of 20 ms. The learned distribution $\mathcal{D}$ is estimated by building a representation $f_\theta: X_{train} \rightarrow \mathbb{R}$ that minimizes the mean square error cost function $C_{\mathcal{D}}(\theta;x_{i})$ parameterized by $\theta$:

\begin{align}
\begin{split}
    \theta^*  & =  \argmin_{\theta} \sum_{x_{i} \in X^{train}} C_{\mathcal{D}}(\theta;x_{i}) = \argmin_{\theta} \sum_{x_{i} \in X^{train}} ||f_\theta(x_{i}) - x_{i}||^2.
\end{split}
\label{eq1}
\end{align}

Let assume the input space of \textbf{X} is $\mathcal{X}$ and the latent space of \textbf{L} is $\mathcal{L}$. Learning a representation using an \emph{Autoencoder} topology consists of mapping the input space $\mathcal{X}$ into $\mathcal{L}$ using the \textbf{Encoder} while mapping it back from $\mathcal{L}$ to $\mathcal{X}$ using the \textbf{Decoder} as shown in figure~\ref{RCAEblock}.

\begin{figure}[!t]
\centerline{\includegraphics[width= 2.5 in]{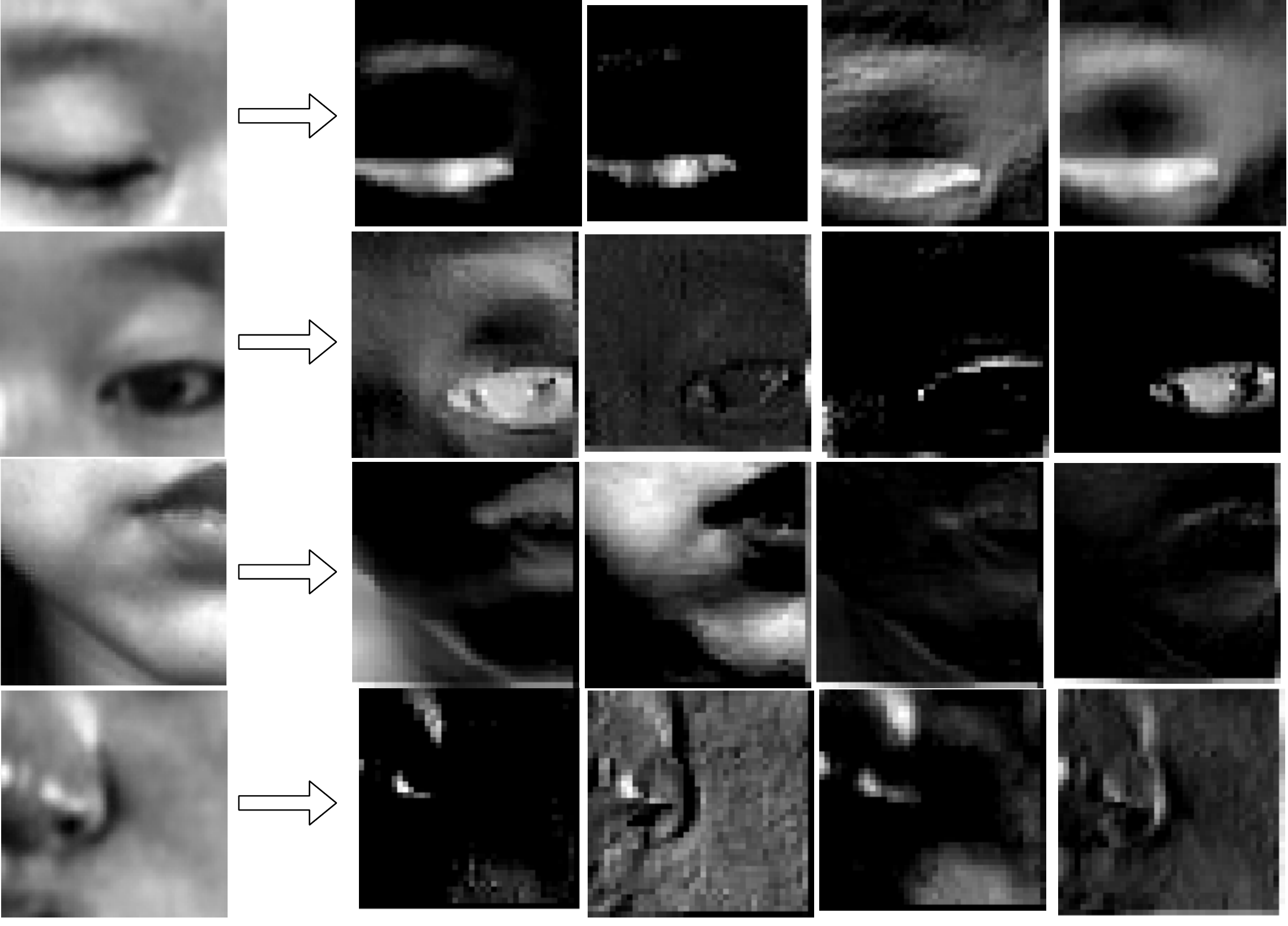}}
\caption{Filters responses at convolutional layer number 2.}
\label{FeatureMapsLayer4}
\end{figure}

\noindent \textbf{Recurrent Convolutional AutoEncoder}. Figure \ref{RCAEblock} illustrates the architecture for capturing visual motion features. The \textbf{Encoder} is designed by stacking four multiple layers of convolution, where each layer is followed by a stride of 2 for down sampling. During convolutional layers, multiple input activations within a filter window are fused to output a single activation. Those CNNs normally extract from each block spatial features related to the spatial patterns. They learn special filters for capturing angles, deformations, edges and other types of appearance features and textures. They encode the primary components of the facial blocks. Some of the feature map activations are represented in figure~\ref{FeatureMapsLayer4}, which shows the presence of specific visual features or patterns that are informative and less redundant compared to the original image patch. For example, the top row of figure~\ref{FeatureMapsLayer4} represents the response to the eyebrows and closed eye patterns, same for the lips corner and the nose in the last two rows.

The feature maps of the final layer Conv4 are fed to a ConvLSTMs module. ConvLSTMs module aim at modeling the spatiotemporal information of short and long motions and providing a feature vector that encodes both spatial appearance and motion patterns within a sequence. This module is designed by stacking two layers of ConvLSTMs, wherein, the cell state of ConvLSTM-1 is fed to the ConvLSTM-2, to preserve the previous spatiotemporal information. Once the encoder reads all the input sequences, it produces the latent representation that encodes the spatiotemporal information of the block sequence. The latent representation \textbf{L} for each time step, is formed by concatenating the outputs (hidden states) of ConvLSTM-1 and ConvLSTM-2. 

A \textbf{Decoder} is designed to map back the latent representation \textbf{L} onto its original space. First it starts by reading \textbf{L} where, $L \in \{L_{1},...,L_{20}\}$, then it decodes the visual motion vector within the $L$ through ConvLSTM-3 and ConvLSTM-4 and finally it outputs a sequence of vectors. The output is followed by four deconvolutional layers with stride by 2 for up sampling. Then, a resizing layer is designed using the nearest neighbours to preserve the spatial structure as the original input. During deconvolutional layers, a single input activation with multiple outputs is obtained. Deconvolutional layers decode the full spatial features using the feature maps of the previous layer in order to recover the details of the facial block components and usually they are considered as learnable up-sampling layers. The feature maps of Deconv4 are fed into a final convolutional layer that maps them into a single input channel with a linear activation function, in order to obtain the logits or the reconstruction of the frames. The network parameters are optimized by minimizing the mean square error function between the reconstructed features and the input features of the entire input sequence (equation~\eqref{eq1}) using the Adaptive Moment Estimation~(ADAM) \cite{kingma2014adam}. 

Due to the complexity of the deep recurrent autoencoder, training and scalability become an issue, bringing poor generalization. It has been proved in \cite{hinton2006fast} that a gradient-based optimization starting with a random initialization appears to often get stuck in poor solutions specially for deep architecture. Thus, in \cite{hinton2006fast} a greedy layer-wise training strategy has been proposed, bringing better generalization and helping to mitigate the difficult optimization problem of deep networks by better initializing the weights of all layers. The layer-wise training works by training one layer at a time. The subsequent layer is then stacked at the top of the features produced by the previous layer and the whole model is retrained again. In our proposal, we use a layer-wise training strategy by first training the convolutional autoencoder. Then we modify the convolutional autoencoder architecture to include the recurrent layers (ConvLSTMs) while loading back the trained weights as a way for re-initializing the network parameters. 

The number of filter is set to 128 and the filters size is set to $3 \times 3$ in all convolutional and deconvolutional layers. The considered activation function is the rectified linear unit function. Each layer is followed by a layer normalization that normalizes the activity of the neurons. For the recurrent part of the encoder and the decoder, 64 filters are considered with size $3 \times 3$ and the hyperbolic tangent function is used as an activation function. A dropout \cite{srivastava2014dropout} with a probability of $65\%$ is applied on the cell states as a regularization technique to reduce overfitting and to enhance generalization. 

\noindent \textbf{Formal Representation}. Let us first simplify the process of building the \textbf{convolutional autoencoder} then the process of building the \textbf{recurrent convolutional autoencoder} . Given an input sequence (instance from bag $i$) \textbf{$x_{i}$} = $(x_{1},...,x_{t},...x_{20})$ such that, \textbf{$x_{t}$} $\in \mathbb{R}^{1\times90\times90 \times1}$. Let \textbf{$x_{t}$} $\equiv h_{0}$ be a single channel input of spatial size $90\times90$. Each convolutional layer \textit{l} maps the previous input at layer \textit{l-1}, into a set of feature map \textbf{$h_{l}$}. The latent map \textbf{$h_{l}$} obtained by the \textit{k}th filter of layer \textit{l} after the convolutional layer \textit{l} is:

\begin{equation}
    h_{l}^{k}= \text{ReLU}(h_{l-1}^{k} \ast W_{l}^{k} + b_{l-1}^{k}),
    \label{eq2}
\end{equation}

\noindent where $\text{ReLU}$ is the rectified linear unit function, $\text{ReLU}(x) = \max(0,\text{x})$, which is mostly used with the convolutional operation ($\ast$). \textit{$W^{k}$} are the weights of the \textit{k}th filter, and \textit{$b^{k}$} is the bias of the \textit{k}-th feature map of the current layer. The latent map $h_{l}$ is mapped back into its original space using a deconvolutional operation, resulting $\reallywidehat{\textbf{x}}$. Equation \eqref{eq1} is the objective function that is used to minimize the reconstruction error and update the parameters $\theta = \{W,b\}$.

To extend the convolutional autoencoder into a \textbf{recurrent convolutional autoencoder}, the latent space $h_{l}$, which is the latent map obtained at the last convolutional layer \textit{l} is fed into a ConvLSTM module. 


\begin{figure}[!t]
\centering
\begin{minipage}[b]{.48\textwidth}
  \centering
  \includegraphics[width=1.\linewidth]{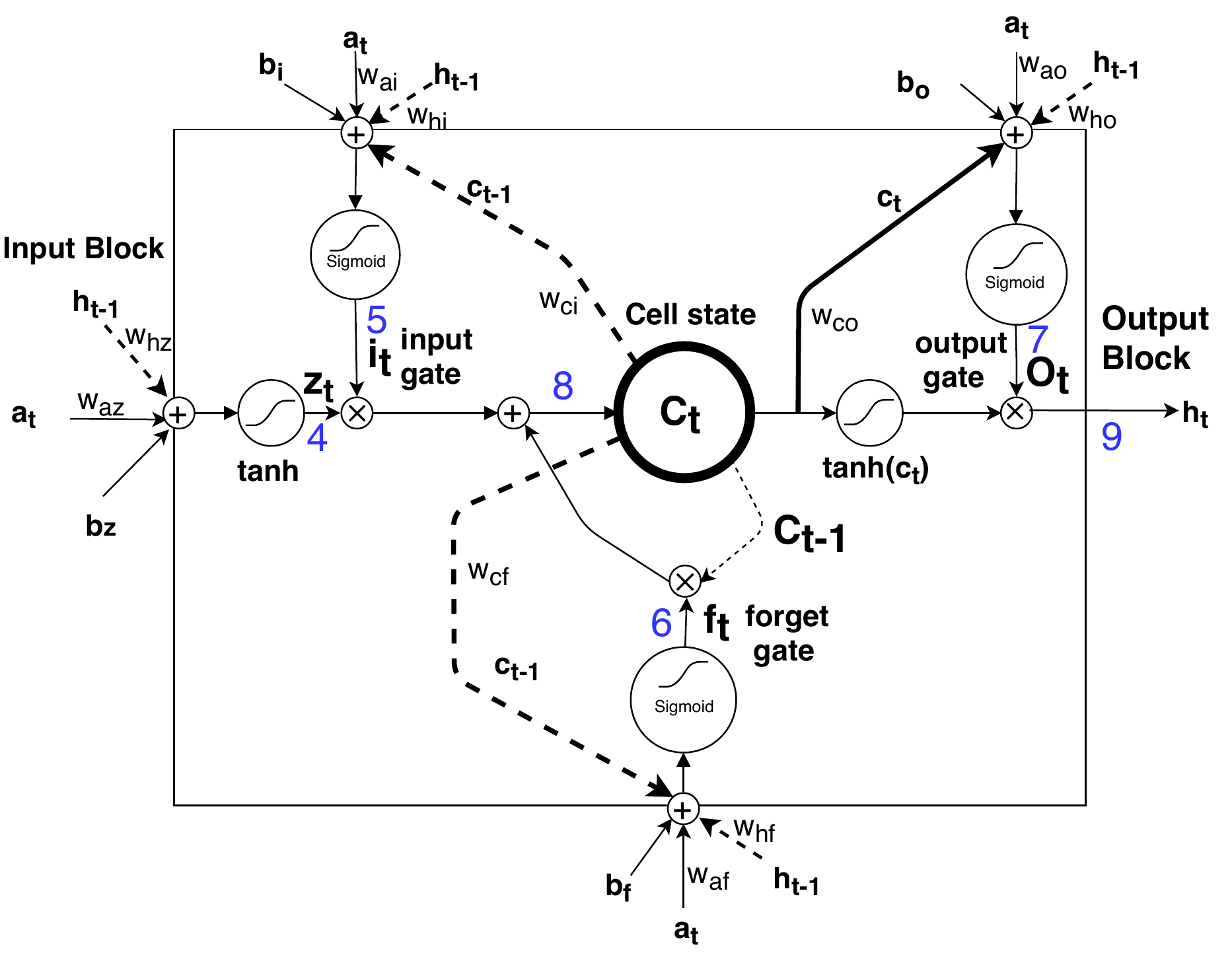}
  \captionof{figure}{Convolutional Long Short-Term Memory Cell.}
  \label{LSTMCellstate}
\end{minipage}%
 \hspace{0.02\textwidth}
\begin{minipage}[b]{.48\textwidth}
  \centering
  \includegraphics[width=1.\linewidth]{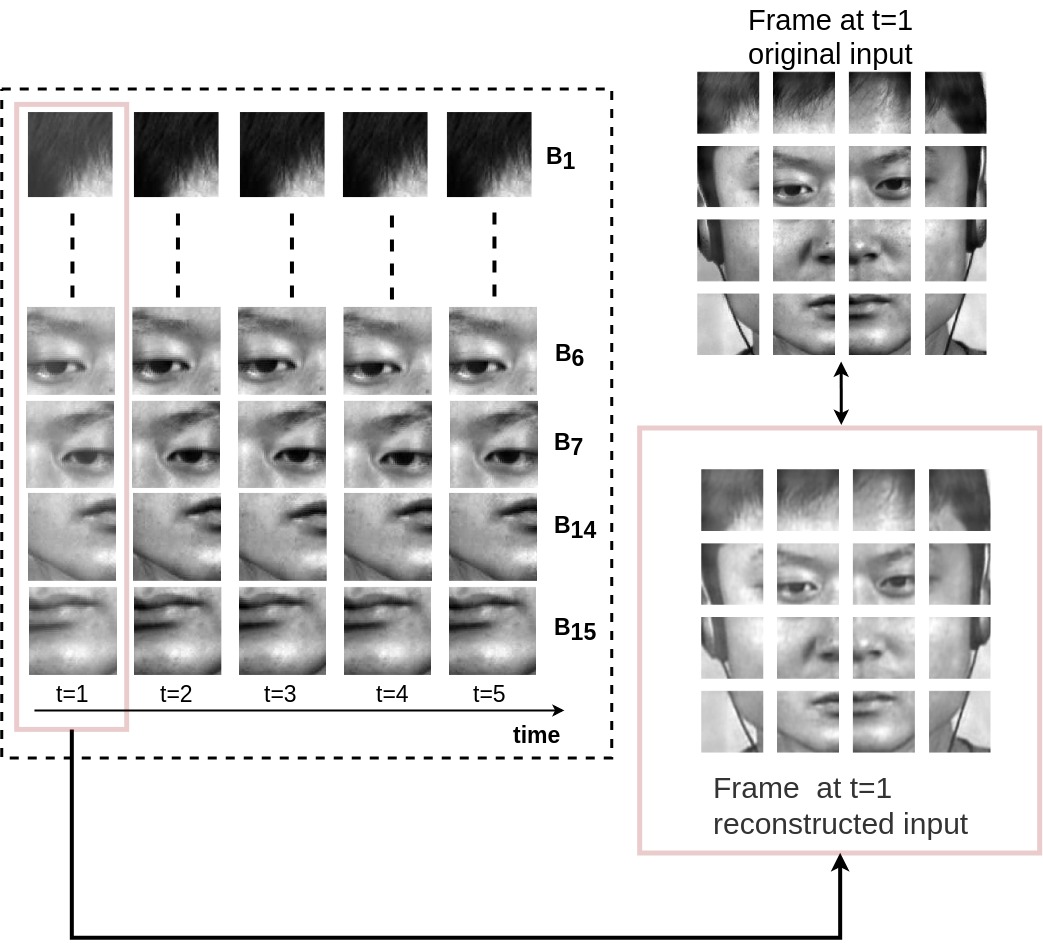}
  \captionof{figure}{The reconstructed blocks over the first 5 time instances.}
  \label{ReconstructedSamples}
\end{minipage}
\end{figure}

Let $h_{l=4}$, represented as \textbf{a}, be the feature map obtained by the layer number 4 from the encoder layer, where $\textbf{a} \in \mathbb{R}^{20\times6\times6\times128}$ as shown in figure~\ref{RCAEblock}. The recurrent neural network needs first to compute the hidden vector sequence $h^{rnn} =\{h^{rnn}_{1},...,h^{rnn}_{t},...,h^{rnn}_{20}\}$ solved through an iterative process: 
\begin{equation}
    h^{rnn}_{t} = \tanh(W_{ih}a_{t} + W_{hh}h^{rnn}_{t-1} + b_{h}),
    \label{eq3}
\end{equation}

\noindent where $W_{ih}$ and $W_{hh}$ denote the input-hidden and hidden-hidden weighting matrices while $b_{h}$ is the bias vector. The hyperbolic tangent function is denoted as $\tanh$. Usually it is preferable over$\text{ReLU}$ in recurrent layer since it is bounded and prevents gradient descent vanishing and exploding phenomena, because its second derivative can sustain for a long range before going to zero.

ConvLSTMs use 2D-grid convolutions to leverage the spatial correlations of input data. Its convolutional structures in both the input-to-state and state-to-state transitions model the spatiotemporal links quite well. Formally, the inputs, the cell states, the hidden states and the gates of ConvLSTM are 4D tensors whose first dimension represents the time step, the second and the third are the spatial dimensions (height and width), and the last dimension is the feature map. The weight matrices here represent the 2d-convolutional kernels. The computation of the hidden value $h_{t}$ of a Conv-LSTM cell is updated at every time step $t$. Figure~\ref{LSTMCellstate} shows a diagram which explain how ConvLSTM unit operates. As shown, the inputs coming from different sources get convoluted with their filters, added up along with bias. It operates by learning gates functions that determine whether an input is significant enough to be memorized, to be forgotten or to be sent to the output. By using a gated way for sorting information over short or long time ranges, the discriminant spatiotemporal information is extracted.

\noindent \textbf{ConvLSTM Formal Representation}: Let \enquote{$\ast$}, \enquote{$\otimes$} and \enquote{$\sigma$} represent respectively the convolutional operation, the Hadamard product and the sigmoid function. The ConvLSTM is formulated as:

\begin{gather}
        \textbf{z}_{t} = \tanh(\textbf{a}_{t} \ast W_{az} + \textbf{h}_{t-1} \ast W_{hz}+ \textbf{b}_{z})  \\ 
        \textbf{i}_{t} = \sigma_{i}(W_{ai}\ast\textbf{a}_{t}+W_{hi}\ast\textbf{h}_{t-1}+W_{ci}\otimes\textbf{c}_{t-1}+\textbf{b}_{i}) \\ 
        \textbf{f}_{t} = \sigma_{f}(W_{af}\ast\textbf{a}_{t}+W_{hf}\ast\textbf{h}_{t-1}+W_{cf}\otimes\textbf{c}_{t-1}+\textbf{b}_{f})\\ 
        \textbf{o}_{t} = \sigma_{o}(W_{ao}\ast\textbf{a}_{t}+W_{ho}\ast\textbf{h}_{t-1}+W_{co}\otimes\textbf{c}_{t}+\textbf{b}_{o})\\ 
        \textbf{c}_{t} = \textbf{z}_{t} \otimes i_{t} + \textbf{c}_{t-1} \otimes \textbf{f}_{t} \\ 
        \textbf{h}_{t} = \tanh(\textbf{c}_{t}) \otimes \textbf{o}_{t},
        \label{Eq8}
\end{gather}

\noindent where \textcolor{black}{\textit{z}}, \textcolor{black}{\textit{i}}, \textcolor{black}{\textit{f}}, \textcolor{black}{\textit{o}} and \textcolor{black}{\textit{c}} are respectively the input block,input gate, forget gate, output gate and cell activation 4D tensors, all having the same size as the tensor $h_{t}$ defining the hidden value. The term $a_{t}$ is the input of a memory cell layer at time $t$. $W_{az}, W_{ai}, W_{hi}, W_{af}, W_{hf}, W_{cf}, W_{ac}, W_{hc}, W_{ao}, W_{ho}$ and $W_{co}$ are the weight matrices, with subscripts representing from-to relationships. For example, $W_{ai}$ being the input-input gate matrix connecting $a_{t}$ to $i_{t}$ as shown in figure~\ref{LSTMCellstate}, while $W_{hi}$ is the hidden-input gate matrix, and so on. The bias vectors are: $bz, bi, bf , bc$ and $bo$. The layers' notation has been omitted for clarity. The activation of the ConvLSTM units is calculated as for the RNN represented in equation~\eqref{eq3}. \textcolor{black}{By referring to equation~\eqref{Eq8}, the current cell state is updated based on the current filtered input added up with its previous filtered cell state. Therefore, the current cell state aggregates useful space-motion information sequentially along the time input space. In such a way, the last state is considered as the final representation that summarizes the contextual information of the entire sequence.}

\noindent \textbf{Qualitative Analysis}. In order to assist the performance of the recurrent convolutional autoencoder on its ability for learning good representative features, the reconstructed blocks over time are drawn for unseen test samples from \textsc{casme-ii} database. Figure~\ref{ReconstructedSamples} demonstrates the effectiveness of the learned model at capturing effective spatiotemporal patterns retaining information about appearance and motion. This information allows a good reconstruction of the input sequence. Our demonstration shows that the learned model is not learning the identity map. Instead, it is capable at extracting motion and visual information present in unseen samples. In addition, being able to decode back to the original form, which requires decoding the spatiotemporal information stored in the latent representation.

\subsubsection{Modeling the Normal Facial Behavior Distribution}


\noindent Having at disposal efficient spatiotemporal features that correspond to the latent space encoded from RCAE for each bag of NFB events, it is possible to model the distribution of these bags separately to establish bags' PDFs. To model the distribution of each bag, a GMM is utilized. The choice for GMM is due to its ability at learning hidden structures within the latent manifold and for its high detection accuracy. 

\begin{figure}[!t]
\centerline{\includegraphics[width=3.5 in]{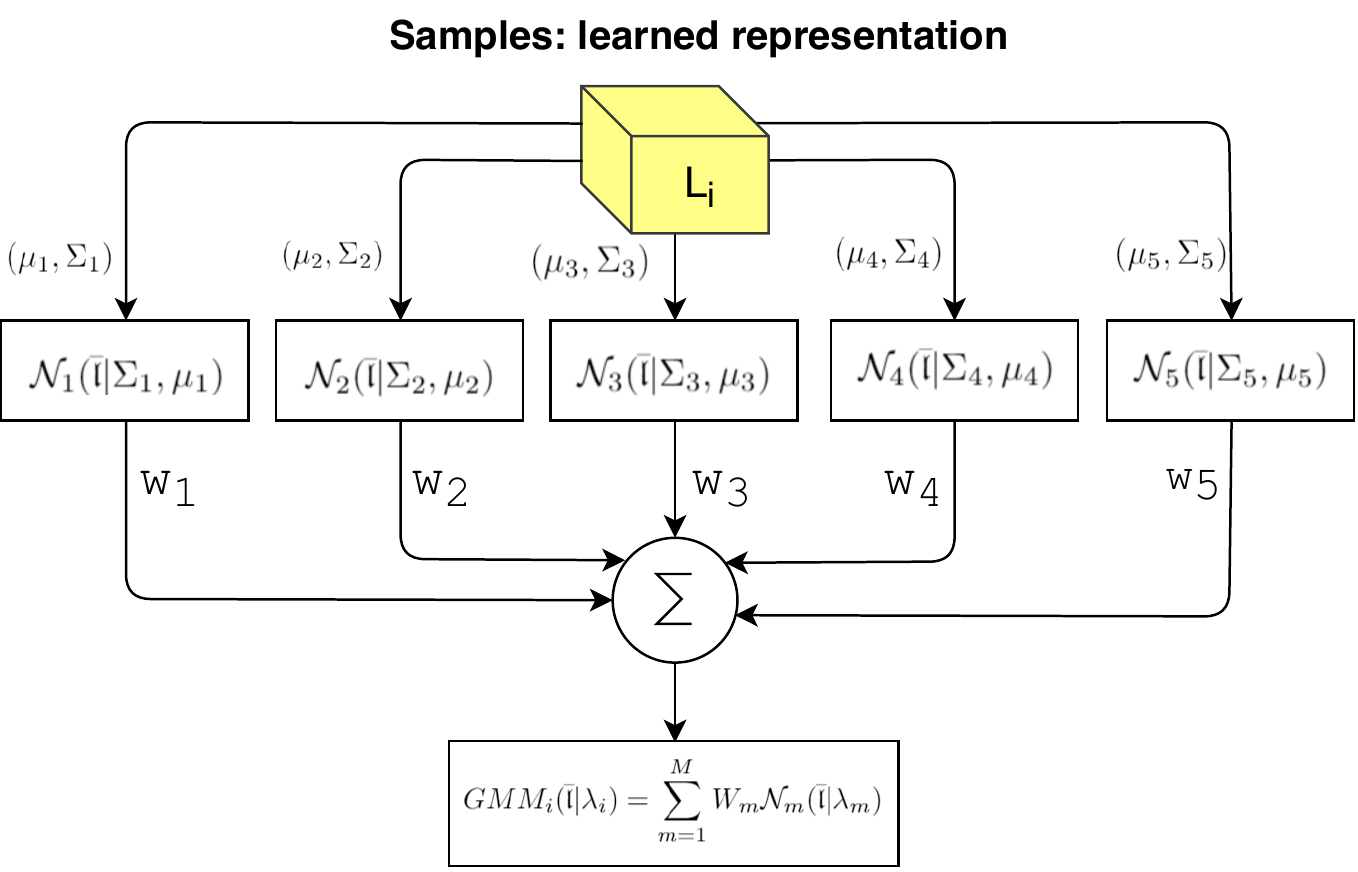}}
\caption{GMM: weighted sum of 5 component densities.}
\label{GMM}
\end{figure}

\noindent \textbf{Gaussian Mixture Model}. Let us consider a set of training samples $\textbf{X}^{train}_{i}$, where $x_{i} \in \mathbb{R}^{20 \times 90 \times 90 \times 1}$ is the sample coming from the facial sub-region bag \textit{i} of 20 time steps. Let us assume $L^{train}_{i}$ is the latent space represented by \textsc{rcae} for a bag $i$, whose instances at time $t$ will be denoted by $\ell_{i,t}\in\mathbb{R}^{1\times6\times6\times128}$, resulting a feature vector of dimension 4608. To simplify the notation, the dependence on $i$, $t$ will be omitted in the rest of the paper, turning it into $\ell$. 
In order to allow a good spatial localization with a low number of mixture components, $\text{\textsc{gmm}}_{i}$ models, $i \in \{1,...,16\}$, are learned separately from $L^{train}_{i}$. Otherwise, it is possible to learn a single \textsc{gmm} over the entire blocks but with a larger number of mixture components, the complexity and the model convergence become an issue.

\noindent An overview of the $GMM_{i}$ structure is represented in figure~\ref{GMM}. The weighted sum of the \textit{m} component densities, \textit{m}=$\{1,...,5\}$, is $GMM_{i}$. Each $GMM_{i}$ model is parameterized by $\lambda_{i} = \{W_{i},\mu_i, \Sigma_i\}$ from all component densities, where $\mu_i$=$\{\mu_1,...,\mu_m\}$ is the mean vector, and $\Sigma_i$=$\{\Sigma_1,...,\Sigma_m\}$ is the covariance matrix and $W_{i}$=$\{W_{1},...,W_{m}\}$ is the weight matrix.

\begin{equation}
    \text{GMM}_{i}(\ell|\lambda_{i}) = \sum_{m=1}^{M} W_{m} \mathcal{N}_{m}(\ell|\lambda_{m}),
    \label{eqWeightedProb}
\end{equation}

\noindent such that $\sum_{m=1}^{M} W_{m} = 1$. The Gaussian distribution of one component density \textit{m} is given by:

\begin{equation}
 \mathcal{N}_{m}(\ell|\lambda_{m}) = \frac{1}{(2\pi)^{d/2}}\frac{1}{|\Sigma_{m}|^{1/2}}exp\{-\frac{1}{2}(\ell-\mu_{m})^T\Sigma_{m}^{-1}(\ell-\mu_{m})\}.
\end{equation}


\noindent One covariance matrix for each Gaussian component is defined and the full rank covariance matrices for the models are estimated. The main goal is to estimate the parameters $\lambda_{i}$ of each $\text{GMM}_{i}$, which matches the best distribution of training feature vectors $\ell$. Given a parameter set initialized by K-Means algorithm, the Expectation-Maximization algorithm estimates the optimal parameter set $\lambda_i$ that maximizes the average likelihood of the training set. In order to determine the suitable number of components \textit{M}, mixture models with different numbers of components are tested, mainly M=$\{2,5,10\}$ over the validation set.

\subsection{The Inference Stage}
\label{inferencestage}

 \noindent At inference stage, as shown in figure \ref{generalblockdiagram}, the full video clip including NFBs and MEs frames is presented. The spatiotemporal features for each block are encoded and fed to the learned GMM to rank the output over a grid of spatial blocks. By thresholding the output, the temporal region and the spatial blocks related to MEs are spotted.

\subsubsection{Output Ranking}
\noindent NFB observations are statistically modeled by a joint probability. Such a PDF captures the correlations between the features and produces the data likelihood for a particular NFB observation, assuming that it is normal and independent of previous observations (independent and identically distributed). The distribution is estimated using a large dataset that has been generated from the PDF we seek for, for instance for NFB events. Once the PDF for NFB is modeled, it is possible to compute the Bayesian posterior probability that a facial behavior observation is normal. But, the PDF for abnormal facial events such as MEs is unavailable, since we exclude it from the entire process due to the lack of data related to MEs. Otherwise, it would be possible to model both distributions and distinguish an abnormal behaviour from a normal one using the Bayesian posterior probability as an activation signal. Nonetheless, in the presence of the NFB distribution only, the weighted log likelihood (equation~\eqref{negativeloglike}) of any observation is an indication of the degree to which the corresponding observation is normal. For example, if the weighted log likelihood is below a particular threshold, it is very unlikely that it was generated from the normal facial behavior PDF and thus it is most probably caused by an abnormal facial behavior event.

\subsubsection{Adaptive Thresholding for Decision Making}

\begin{figure*}[!t]
\centerline{\includegraphics[width= 6 in]{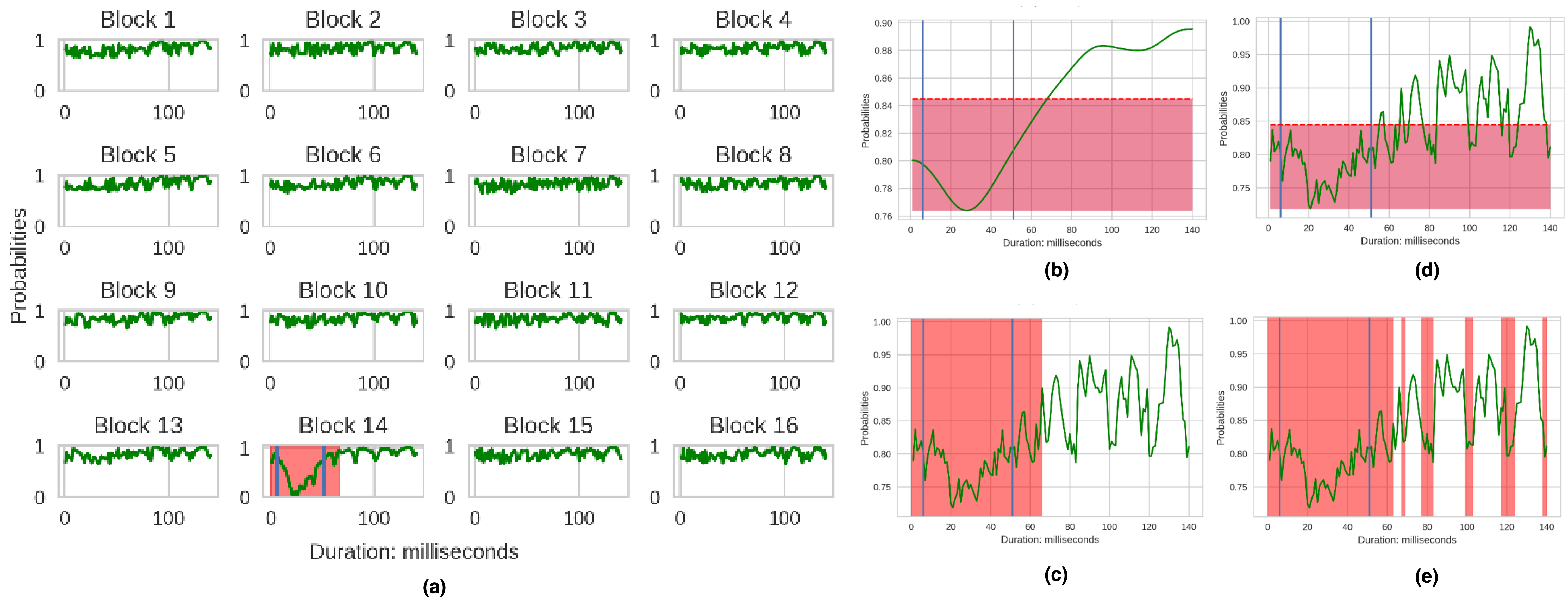}}
\caption{Decision making process via adaptive thresholding.}
\label{Localization}
\end{figure*}

\noindent Decision making is the process of identifying temporal bounds and spatial location of MEs in a facial video clip. The process is based on the weighted log likelihood as a score which is thresholded in an adaptive way. A general representation of the thresholding process is demonstrated in figure~\ref{Localization}. Firstly the temporal bounds of any ME are spotted and then the spatial occurrence of appearance changes are detected. Let $P_{block}(\ell)$ be the weighted log likelihood at time \textit{t} of block $B_{i}$ as represented in figure~\ref{Localization}(a), 
$P_{block}(\ell)$ is computed using equation~\eqref{negativeloglike}. 

\begin{equation}
    P_{block}(\ell) = log(\sum_{m=1}^{M} W_{m}\times\mathcal{N}_{m}(\ell|\lambda_{m})).
    \label{negativeloglike}
\end{equation}

In order to describe the spatio-temporal changes of the video in a compact and consistent representation, we perform mean pooling across time over $P_{block}(\ell)$ to obtain a composite curve as shown in figure~\ref{Localization}(c) and it is referred to as $P_{video}$. $P_{video}$ is computed using equation \eqref{eqavgprobabilities} and it represents the average weighted log likelihood of the video. The intuition behind this pooling step is to help in accentuating the appearance-motions changes by aggregation of these local weighted log-likelihood.

\begin{equation}
    P_{video}= \frac{1}{16}\sum_{i=1}^{16} P_{block,i}.
    \label{eqavgprobabilities}
\end{equation}

By applying a 1D-Gaussian kernel over $P_{video}$ with a standard deviation $\sigma=10$, a smoother curve is obtained as shown in figure~\ref{Localization}(b), and we refer to as $P_{sv}$. Smoothing is required in order to distinguish relevant peaks from local magnitude variations and noise. For instance, if a threshold $\mathcal{T}$ is set manually around $0.85$ and applied over $P_{sv}$ (figure~\ref{Localization}(b)), only relevant peaks are picked out as shown in figure~\ref{Localization}(c), which correspond to the fastest facial movements. In contrast, if the same threshold is applied over an unsmoothed graph that corresponds to $P_{video}$ as shown in figure~\ref{Localization}(d), many false positive peaks are detected as shown in figure~\ref{Localization}(e).

For adaptive thresholding, a threshold $\mathcal{T}$ is defined over the smooth graph $P_{sv}$ in order to locate the temporal bounds where MEs occur, as:
 
 \begin{equation}
    \mathcal{T} = \max(P_{sv}) - \mu(P_{sv}) + \min(P_{sv}) + 0.5\times \sigma(P_{sv}).
    \label{eqadaptive}
\end{equation}
 
\noindent where $\mu$ and $\sigma$ are the mean and the standard deviation. Once $\mathcal{T}$ is computed, it is applied over $P_{sv}$ (figure~\ref{Localization}(b)), all likelihoods lower than $\mathcal{T}$ are considered as MEs frames. By that, the temporal bound is obtained. To specify the spatial location of MEs occurrences within the image, $\mathcal{T}$ is applied over each block separately $P_{block}$ (figure~\ref{Localization}(a)). As a result, it appears that only $B_{14}$ which corresponds to the left lip corner is the sub-region associated with a fast motion and facial appearance changes. 

\noindent \textbf{Special Case:} A video clip is presented to ADS-ME framework such that only NFBs are presented and no occurrence of MEs exist, then we experience that at each time instance the computed probability is associated with high score. Afterwords while computing an adaptive threshold $\mathcal{T}$, no temporal or spatial bound is located since as $\mathcal{T}$ is always lower than the minimal probability value.

\section{Experiment, Results and Discussion}
\label{exp}

\noindent To validate and select the best performance of the proposed algorithm among a set of hyper-parameters to control, the following experiments are conducted: regarding the \textit{Feature Learning Stage}, firstly, the effect of TW = $\{10, 20, 30\}$ ms is evaluated. Secondly, the influence of the temporal multiscaling while modeling NFBs by setting S=$\{{1,2,3}\}$ and S=$\{{1}\}$ in two distinct experiments is tested. Thirdly, in order to validate the robustness of the hierarchical spatiotemporal representation for capturing motions and spatial changes using RCAE, a comparison with other existing spatio-temporal feature extraction approaches as LBP-TOP and 3D-HOG is performed. Fourthly, in order to determine the appropriate number of mixture components for the GMM model, the validation set is used to select the best model within M =$\{{2,5,10}\}$. Regarding the \textit{Inference Stage}, the effectiveness of the adaptive thresholding technique for decision making is compared with the obtained results when using cross validation for determining the best threshold. To evaluate and compare the performance of ADS-ME algorithm with the state-of-the-art methods, experiments are conducted on three benchmarks: \textsc{casme-i} \cite{yan2013casme}, \textsc{casme-ii} \cite{yan2014casme} and \textsc{smic-hs} \cite{li2013spontaneous}. 

\subsection{Databases and Training Protocol}

\noindent \textsc{casme-i} database: It contains 195 spontaneous ME sequences that correspond to 19 subjects. It is recorded at 60 fps. It is divided into two sets CASME-A (subjects 1 up to 7) and CASME-B (subjects 8 up to 19). CASME-A is recorded under natural light conditions and with a high spatial resolution ($1280 \times 720$) resulting about 280$\times$340 pixels on the facial area. While CASME-B is recorded in a room under 2 led lights with a lower resolution camera ($640 \times 480$) resulting about $150\times190$ pixels on facial area. Out of 195 sessions around 172 sessions are valid for spotting MEs, with the ground truth available for the onset and the offset frames.

\noindent \textsc{casme-ii} database: It is recorded under a \textit{high temporal resolution} (200 frames per second (fps)) and a \textit{spatial resolution} of 280$\times$340 pixels on facial area. The total number of sessions is 255  that correspond to 26 subjects. Each session is a short video clip up to few seconds. With our method of block division and random sampling with three temporal multiscaling windows, $3\times16\times n$ samples instances are obtained.

\noindent \textsc{smic-hs} database: It contains 160 spontaneous ME sequences that correspond to 16 subjects. Clips from all participants were recorded with a high speed camera at 100 fps with standard spatial resolution $640\times480$, resulting about $150\times190$ pixels on the facial area.

\noindent \textbf{\textit{Training protocol setting across the databases}}: The precision of the proposed detection method is evaluated in a \emph{Subject-Independent} manner.

\begin{itemize}
    \item \textsc{casme-i}: for the \textit{validation set}: subjects $\{2,9,13\}$ are considered with 18 sessions while for the \textit{test set}: subjects $\{6,12,18\}$ are considered with 28 sessions. The rest of the subjects are considered as the \textit{training set} with 126 sessions. 

\item \textsc{casme-ii}: for the \textit{validation set}: subjects $\{23,24, 26\}$ are considered with 39 sessions while for the \textit{test set}: subjects $\{1,7,21,22,25\}$ are considered with 29 sessions. The rest of the subjects are used as the \textit{training set} with 187 sessions. 

\item \textsc{smic-hs}: for the \textit{validation set}: subjects $\{1,4,11\}$ are considered with 29 sessions while for the \textit{test set}: subjects $\{2,8,112\}$ are considered with 30 sessions. For the \textit{training set}, 101 sessions are considered. 
\end{itemize}

The choice for the validation and the test sets are based on providing a variety of subject facial deliberate actions such as head movements and eye blinks, wherein the ability of ADS-ME algorithm can be tested to report to what extend NFBs could be confused with MEs.

\subsection{Evaluation Metrics}

\noindent The Precision, Recall and the area under the ROC curve (AUC) are reported. Moreover, in order to evaluate the temporal boundary precision with respect to the ground truth, the \textit{Mean Average Duration} (\textbf{MAD}) of ME segments, which is the minimal average length of ME segments for a specific database, is estimated and compared with the \textit{Mean Average Shift} (\textbf{MAS}), which is an indicator of how much the prediction of the temporal location is expanded or collapsed compared to MAD. It is defined as:

\begin{align}
        &MAS(q) = \frac{1}{n}\sum_{k=1}^{k=n}|q_{p}-q_{g}|  \\ 
        &MAS(u) = \frac{1}{n}\sum_{k=1}^{k=n}|u_{p}-u_{g}| \\ 
        &MAS = \frac{1}{2} \left(MAS(q) + MAS(u)\right),
\end{align}

\noindent where, $q_{p}$ and $q_{g}$ indicate the predicted and the ground truth of the onset frames respectively. Same for the offset frames represented as $u_{p}$ and $u_{g}$ respectively.

\subsection{Parameters Evaluation and Discussion}

\begin{figure}[t!]
    \centering 
    \includegraphics[width= 3.5 in]{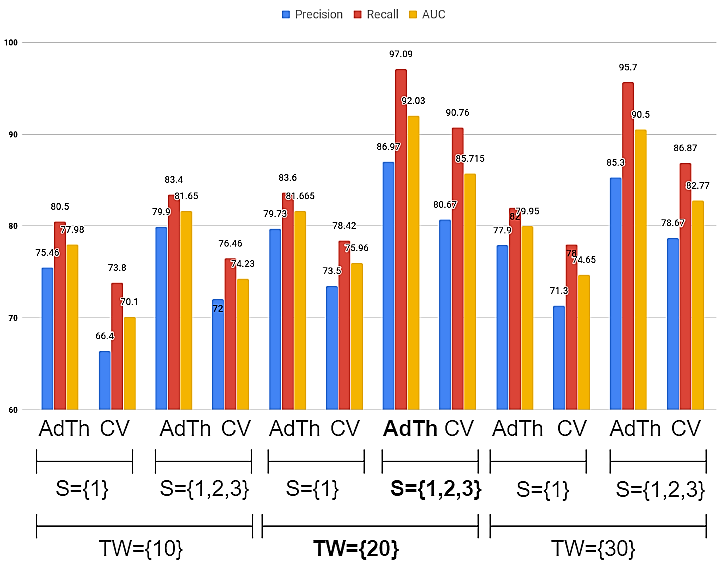}
    \caption{Studying the effects of different time windows \textit{TW}$=\{10,20,30\}$, temporal Multiscaling \textit{S}$=\{1,2,3\}$, and thresholding technique (Adaptive:\textit{AdTh} or Cross Validation:\textit{CV}) over \textsc{casme-i} database with \textit{M}$=\{5\}$.}
    \label{casme1metrics2}
\end{figure}

\noindent For a fair evaluation of the proposed techniques, a control experiment is performed over the \textsc{casme-i} database. It aims at: figuring out the best parameter for \textit{TW}$=\{10~\text{ms},20~\text{ms},30~\text{ms}\}$, evaluating the effect of temporal multiscaling (with and without) and comparing the thresholding techniques: adaptive versus using a fixed threshold obtained by cross validation. CASME-I is considered since it is more challenging than \textsc{smic-hs} and \textsc{casme-ii} due to its low spatial and temporal resolutions. The best parameters obtained using the control experiment are then also used for \textsc{smic-hs} and \textsc{casme-ii} processing.

Nonetheless, to pick out the best number of mixture components $\textit{M}=\{2,5,10\}$, we use the validation set. We find out that when $\textit{M}=\{10\}$ and $\textit{M}=\{5\}$, better results are obtained than with $\textit{M}=\{2\}$. Although the best results are obtained with $\textit{M}=\{10\}$, in order to reduce the complexity of the algorithm, we chose $\textit{M}=\{5\}$, since to some degrees it is still near the best results. In the following we only report the evaluation of the results while $\textit{M}=\{5\}$. The experimental setup for evaluation is performed as demonstrated in figure~\ref{casme1metrics2}.

For each temporal window TW=$\{10,20,30\}$~ms, four experimental results are reported as shown in figure~\ref{casme1metrics2} where each experiment reports Precision, Recall and AUC values. In the first two experiments with \textit{S}=$\{1\}$, no temporal multiscaling is considered, contrary to the third and the fourth experiments with \textit{S}=$\{1,2,3\}$. For each experiment either adaptive threshold \textbf{(AdTh)} or fixed threshold obtained via typical cross validation \textbf{(CV)} evaluation is done. The best obtained results are those under the following experimental setup: \textbf{TW}=$\{20\}$, \textbf{S}=$\{1,2,3\}$, \textbf{AdTh}, which are marked in bold in figure~\ref{casme1metrics2}, while reporting a Precision = 86.97\%, a Recall = 97.09\% and an AUC value = 92.03\%. 
 
For the same parameters setting: S=$\{1,2,3\}$, TW=$\{20\}$, but with the cross validation method to pick out the best threshold $\mathcal{T}$, found at a probability = 0.72, a Precision = 80.67\%, a Recall = 90.76\% and an AUC value = 85.71\% are reported. Obviously, the adaptive thresholding provides better precision and results than with cross validation. Moreover, by comparing the performance with parameters: TW=$\{20\}$ with adaptive thresholding (AdTh) while $S=\{1\}$, a Precision = 79.73\%, a Recall = 83.6\% and an AUC value = 81.66\% are reported. As a result, the temporal multiscaling improves the precision rate from 79.73\% to 86.97\%. This confirms the assumption that the temporal multiscaling enriches the modelling of NFBs and reduces the false positive rate.

\subsection{Detection Results over ME Databases}

\noindent The performances of our algorithm using the best parameters setting are reported with time window for TW=$\{20\}$, temporal multiscaling S=$\{1,2,3\}$, adaptive threshold and M=$\{5\}$ over \textsc{casme-i}, \textsc{casme-ii} and \textsc{smic-hs} databases. Moreover, we report the mean average shift duration in milliseconds to estimate to what extend the estimation for the onset-offset segment frames expand or collapse from the ground truth. For this purpose, the mean average duration (MAD) of MEs over each database is computed and compared with MAS. The $\pm$ sign corresponds to the standard deviation of the sample mean distribution.

Table \ref{table1} shows that spotting MEs on high speed camera and high spatial resolution video sequences improve the precision rate, as demonstrated using the \textsc{casme-ii} database. However, the lower the spatiotemporal resolution, the more challenging MEs spotting is (as the case of CASME-I database) since the motion and the feature displacement become very similar to those with stationary motion. Indeed, developing an accurate ME detection system requires high quality data. In addition, Table \ref{table1} shows that the detection algorithm has an acceptable mean average shift duration compared to the mean average duration of the MEs. This is an important aspect when considering using the detected frames for MEs recognition in a second step.

The proposed method is then compared  with the state-of-the-art methods. The results are presented in Table \ref{table2}. The performances of our algorithm are higher than those in \cite{li2017towards} and \cite{borza2017real}. However, compared to \cite{duque2018micro}, the same performance is achieved over \textsc{casme-ii} database, while having a better score over \textsc{smic-hs} database. It worths to note that state-of-the-art methods focus only on high quality databases (\textsc{smic-hs} and \textsc{casme-ii}). Even though some papers reported acceptable precision rates, those studies forbid NFBs.

\begin{table}[!t]
\centering
\caption{Performance evaluation over the three micro facial expressions databases.}
\label{table1}
\begin{tabular}{@{}llllll@{}}
\toprule
\textbf{Database} & \textbf{Precision} & \textbf{Recall} & \textbf{AUC} & \textbf{MAS} & \textbf{MAD} \\ \midrule
\begin{tabular}[c]{@{}l@{}}\textsc{casme-i} (60 fps)\end{tabular} & 86.97\% & 97.09\% & 92.03\% & 5.1 $\pm$ 0.62 ms & 20 ms \\
\begin{tabular}[c]{@{}l@{}}\textsc{smic-hs} (100 fps)\end{tabular} & 88.87\% & 97.82\% & 93.76\% & 5.9 $\pm$ 0.82 ms & 34 ms \\
\begin{tabular}[c]{@{}l@{}}\textsc{casme-ii} (200 fps)\end{tabular} & 91.6\% & 98.9\% & 95.17\% & 9 $\pm$ 0.95 ms & 66 ms \\ \bottomrule
\end{tabular}
\end{table}

\begin{table}[!t]
\centering
\caption{Reported AUC values in state-of-the-art.}
\label{table2}
\begin{tabular}{@{}lcc@{}}
\toprule
State-of-the-art & \textsc{smic-hs} & \textsc{casme-ii} \\ \midrule
Li et al. \cite{li2013spontaneous} & 65.55\% & NR \\
Liong et al. \cite{liong2014subtle} & 72.87\% & NR \\
Li et al. \cite{li2017towards} & 83.32\% & 92.98\% \\
D. Borza et al. \cite{borza2017real} & NR & 93.4\% \\
C. Duque et al. \cite{duque2018micro} & 89.80\% & 95.13\% \\
\textbf{ADS-ME} & \textbf{93.76\%} & \textbf{95.17\%} \\ \bottomrule
\end{tabular}
\end{table}

\subsection{Feature Learning Strategies and Evaluation}

\noindent For evaluating the efficiency of the proposed feature learning method using RCAE, we compare our method with handcrafted spatiotemporal features, mainly \textsc{lbp-top} and \textsc{3d-hog}. The algorithm architecture remains the same, but in the feature learning stage, the spatiotemporal features over each sample block are extracted using \textsc{lbp-top} and \textsc{3d-hog}. We only report the results using the best sets of parameters: For LBP-TOP : $\text{S}=\{1,2,3\}$, $\text{TW}=\{20\}$, adaptive thresholding and $M=\{7\}$, while for 3D-HOG: $\text{S}=\{1,2,3\}$, $\text{TW}=\{20\}$, adaptive thresholding and $\text{M}=\{4\}$. RCAE has a superior performance over handcrafted features as shown in Table \ref{table3}. Obviously, RCAE is capable at representing and extracting a meaningful spatial and temporal information. Handcrafted descriptors obviously are not able at adapting to represent various speed of motions and spatial displacements.

\begin{table}[!t]
\centering
\caption{Evaluation using different feature representation. AUC values are reported.}
\label{table3}
\begin{tabular}{@{}lccc@{}}
\toprule
Feature extraction & \textsc{casme-i} & \multicolumn{1}{l}{\textsc{smic-hs}} & \textsc{casme-ii} \\ \midrule
LBP-TOP & 66.16\% & 68.3\% & 78.2\% \\
3D-HOG & 61.43\% & 63.87\% & 71.9\% \\
RCAE & \textbf{92.03\%} & \textbf{93.76\%} & \textbf{95.17\%} \\ \bottomrule
\end{tabular}
\end{table}

\section{Conclusion}
\label{conc}

\noindent A novel algorithm to spot MEs spatially and temporally is proposed. The problem of ME detection is reformulated as an anomaly detection problem. Frequent normal facial behaviors are considered as regularities while infrequent facial behaviors such as micro expressions are considered as irregularities. The main strengths of our algorithm are its simplicity and accuracy in detecting MEs while the subjects have been given the freedom to perform any deliberate facial actions while restricting anomalies only to MEs. Moreover, it has reasonable temporal detection deviation and has good detection rate over various spatial and temporal resolution databases. On the contrary, the decoupled learning process between the feature learning and the PDF estimation could be one of the limitations we encounter, which may reduce its efficiency from being end-to-end. Mixture Density Network (MDN) could be a solution, where the RCAE latent space could be fed to MDN to model its distribution alongside spatiotemporal feature extraction. 


\end{document}